\documentclass[12pt,letterpaper]{article}
\usepackage{floatrow}
\newfloatcommand{capbtabbox}{table}[][\FBwidth]

\newcommand{\footremember}[2]{%
    \footnote{#2}
    \newcounter{#1}
    \setcounter{#1}{\value{footnote}}%
}
\newcommand{\footrecall}[1]{%
    \footnotemark[\value{#1}]%
} 

\usepackage{pslatex}
\usepackage{apacite}

\usepackage{kpfonts}
\usepackage{stackengine,scalerel}
\usepackage{calc}
\newlength\shlength
\newcommand\xshlongvec[2][0]{\ThisStyle{\setlength\shlength{#1\LMpt}%
  \stackengine{-5.6\LMpt}{$\SavedStyle#2$}{\smash{$\kern\shlength%
    \stackengine{\dimexpr 1.3pt+6.25\LMpt}{$\SavedStyle\mathchar"017E$}%
      {\rule{\widthof{$\SavedStyle#2$}}{\dimexpr.1pt+.5\LMpt}\kern.4\LMpt}{O}{r}{F}{F}{L}\kern-\shlength$}}%
      {O}{c}{F}{T}{S}}}
      
\usepackage{graphicx}
\usepackage{array} 

\usepackage{caption}
\title{Opacity, Obscurity, and the Geometry of Question-Asking}

\author{
Christina Boyce-Jacino\footremember{sds}{Social and Decision Sciences, Dietrich College, Carnegie Mellon University, 5000 Forbes Avenue, Pittsburgh, PA 15015} \and
Simon DeDeo\footrecall{sds} \footremember{sfi}{Santa Fe Institute, 1399 Hyde Park Road, Santa Fe, NM 87501; {\tt sdedeo@andrew.cmu.edu}.}
}
\date{}

\begin{document}

\maketitle

\begin{abstract}
\noindent Asking questions is a pervasive human activity, but little is understood about what makes them difficult to answer. An analysis of a pair of large databases, of New York Times crosswords and questions from the quiz-show Jeopardy, establishes two orthogonal dimensions of question difficulty: obscurity (the rarity of the answer) and opacity (the indirectness of question cues, operationalized with word2vec). The importance of opacity, and the role of synergistic information in resolving it, suggests that accounts of difficulty in terms of prior expectations captures only a part of the question-asking process. A further regression analysis shows the presence of additional dimensions to question-asking: question complexity, the answer's local network density, cue intersection, and the presence of signal words. Our work shows how question-askers can help their interlocutors by using contextual cues, or, conversely, how a particular kind of unfamiliarity with the domain in question can make it harder for individuals to learn from others. Taken together, these results suggest how Bayesian models of question difficulty can be supplemented by process models and accounts of the heuristics individuals use to navigate conceptual spaces.

\vspace{2mm}
\noindent \emph{Keywords:} Question-asking; information search; computational linguistics, semantic memory

\end{abstract}

\section{Introduction}

We learn from more than just isolated exploration of our environment; we also seek information from others by asking questions. The aim of these questions may be to gather novel information, check our understanding, or to compare our understanding with others. Just as their aims may vary, questions may also be more or less difficult to answer, even when the person being asked knows the answer. It may be that the answer sought is obscure, and difficult to recall. Or, the question-asker might have phrased the question poorly: rephrased, such a question could become easier.

Questions come in many different forms. In the questions we consider in this work, answers are drawn from a structured semantic space. The answer is correct because of what it means in common or contextual usage. A second property of the questions we consider is that they involve retrieval: the answers are words or concepts previously encountered~\shortcite{Bourgin}, which contrasts with questions, such as mathematics problems, whose answers require novel construction. These two properties, semantic structure and retrieval, help delineate, but do not fully define, the set of questions we consider.\footnote{The problem of a general classification of questions is an open one, and we return to it in the discussion.}

We propose that answering questions like these is made difficult by variation along two primary dimensions: obscurity and opacity. In the case of the first dimension, a question can be difficult because its answer is \emph{obscure}: a word, for example, that occurs infrequently in that question-asking context. From a Bayesian perspective, these questions are difficult because the answerer must accumulate a sufficient amount of evidence or insight to overcome the low prior probability that she attributes to the answer~\cite{chater}. In process models of obscurity, a rare answer may have high information-retrieval costs. There, while an activation of conceptual spaces related to the question may bring to mind common words quite easily, rarer ones are not something ``on the tip of one's tongue''.

\begin{table}[!b]
\begin{center} 
\caption{Examples of questions whose answers involve retrieval from a structured answer space. Obscure questions demand people recall words and names encountered infrequently (if at all): ``casters'', or (for many outside South Korea) the golfer Pak Se Ri. Opaque questions mix domains (books that sit near telephones and websites found online; American men's basketball and South Korean women's golf).} 
\label{typesQ_SA} 
\begin{tabular}{l|ll} 

      &  (Clear)  & Opaque \\ \hline
(Common)       &  What cars do large & The Yellow Pages,  \\
&  families tend to buy?  & but on the screen? \\
& (Minivans) & (Google) \\ [0.25cm] \hline 
Obscure & What are the wheels under  & Who is the Michael Jordan \\ 
& a chair that let it roll? & of South Korean golf? \\
& (Casters) & (Pak Se Ri) \\

\end{tabular} 
\end{center} 
\end{table}
These priors can be domain specific. As early as the seventeenth century, Spinoza proposed that memory was organized as a collection of connections within the mind, whose structure is governed by experience: seeing horse tracks, a soldier is reminded of horsemen and war, a farmer, of plows and fields \cite{Spinoza}. This captures the notion that familiarity governs and aids recollection; in the modern era, ideas about the relationship between familiarity, unexpectedness, and domain specificity have been demonstrated in a variety of experiments~\cite{hills2012optimal,hills2015foraging}. The frequency with which a word has been encountered may affect how its mental representation is encoded~\cite{Preston} by, for example, differential activation states \cite{McClelland1981}. Other work suggests that this encoding happens at a network level, or is influenced by how many words have been encoded as similar to the rare word in question \cite{Vitevitch}. In all of these accounts, words that have been encountered less often, have been seen in fewer contexts \cite{Johns}, or have fewer semantic or phonological neighbors \shortcite{Frauen}, may be harder to retrieve. 

Difficulty may also arise from \emph{opacity}: the relationship between question and answer may be indirect. One sign of opacity is that the question can be made easier by a rephrasing it in a way that does not alter the answer (i.e., keeping obscurity constant). In the Spinozan picture, opacity arises from the way in which the question activates networks of associations within which the answerer searches. Say a tongue-tied individual in a hardware store is looking for the aisle with the mops. If he asks about the ``broom, but for when I spill things,'' he refers to the same word (mop) as one who asks about the ``three-foot fluid matter absorbing wand''. The first formulation locates the answerer in the semantic space of domestic objects and events while the second uses less context-consistent terminology more distant from the domain. The individual who poses the first question will be more easily shown to the right aisle. 

Opacity causes difficulties on a day-to-day basis, particularly for newcomers to a task, subject matter, or environment. The child who asks for the ``screwdriver with the cross'' requires the shopkeeper to translate from abstract shapes to tools for home improvement; the student who asks about ``wanting to keep stuff you already have, even when it's silly'' requires a teacher to map from the space of folk psychology to science; a visitor to London who asks for ``downtown'' requires a native to translate a concept associated with American cities (``vibrant place of attractions and street-life'') into the semantic space of locations in London. Table~\ref{typesQ_SA} shows an example of the four question types that arise from the two contrasts of opaque vs clear and obscure vs common.

We take a data science approach to study these two dimensions. We draw on two large corpora, a database of 471,342 \emph{New York Times} crossword puzzles from 1993 to 2017, and set of 104,293 questions from the quiz show \emph{Jeopardy}. All questions in these two data sets satisfy our conditions of having answers which are both retrieved and structured. They also have the particular quality of being highly specific: for all questions in our set, there is only one correct answer. Finally, the two sets also allow us to relate features of a question-answer pair to difficulty, since both assign a level of difficulty to each question. The benefits of such a corpus do not come without limitations: most importantly, these questions are posed by people who already know the answer, and what difficulty the answerer encounters has been consciously built in. We address these limitations further in the discussion.

We operationalize our first dimension, obscurity, using the Google Natural Language Corpus to quantify (context-free) word frequency. We operationalize our second dimension, opacity, using the semantic representations provided by the Machine Learning tool word2vec which enable us to measure the indirectness of a the relationship between question and answer. The measure quantifies the strength of latent cues within a question, and can characterize the extent to which these cues function independently or in synergistic combination. %We can then quantify the extent to which questions point to answers through the intersection of semantic spaces, and how the semantic linkage of questions to answers can be made more or less difficult to navigate.

We find three key results. First, a strong relationship between difficulty and obscurity: easier questions have more common answers. We also find a strong relationship between difficulty and opacity; in the case of crosswords, this effect dominates over obscurity. Obscurity and opacity also trade off: for questions with the same level of difficulty, an increase in obscurity can be compensated for by a reduction in opacity. %Controlling for obscurity and increasing opacity (or the converse) increases difficulty, demonstrating both the independence and compensatory ability of the two dimensions. These results generalize across the two question-asking contexts. 

Second, opacity is itself multidimensional: while difficult questions are in both cases more indirect (as measured by our machine-learning models of semantic content), the cues present in easy questions provide a greater benefit to those who can combine them in a synergistic fashion. More difficult questions employ more distantly-related cues and frustrate this synergistic strategy. 

Opacity and obscurity are not the only dimensions for difficulty in structured-retrieval questions. Our third result demonstrates the existence of additional dimensions of difficulty, including the rarity of the question words themselves, the question's logical and syntactical complexity, and the local density of the semantic network in which the answer lies.

\section{Methods}

The two dimensions we propose and operationalize can be motivated intuitively by examples of the two modes, as shown in Table~\ref{2x2}. Take, for instance, the two ``clear and direct'' clues, i.e., two clues with similar and low levels of opacity. Both clues provide straight forward, dictionary definitions of their answers. The second is harder because of its obscurity: ``kennel'' is relatively uncommon word. In contrast, opacity can make a question harder even when word frequencies are relatively high. The concepts that a question evokes may be distant from the space in which the answer is typically encountered, or they may be scattered over a number of different conceptual domains, such as ``iron horse'' for train, or ``watery knowledge'' for hydrology. Interlocutors encountering an opaque question may consider it enigmatic or even incomprehensible. 

\begin{table}[!t]
\begin{center} 
\caption{Obscurity and Opacity. For each question-answer pair we present the opacity of the question as the angle of cosine similarity (Eq.~\ref{vec_example}; larger angles indicate higher opacity), and the obscurity of the answer in terms of frequency per million words. Clear and direct questions with common answers, such as a ``game with bat and ball'' have low opacity and low obscurity. Difficulty, as we will show, increases both with obscurity and opacity (moving either to the right, or down, or both).} 
\label{2x2} 
\resizebox{\textwidth}{!}{%
  \begin{tabular}{l|ll|ll} 

      &  (Clear) & & Opaque \\ \hline
  (Common)       	&  Game with bat and ball:  &  Baseball & Iron horse:  &  Train\\
  & \emph{Opacity}: $63.4^\circ$ & \emph{Obscurity}: 33.4 wpm & \emph{Opacity}: $77.5^\circ$ & \emph{Obscurity}: 87.1 wpm  \\ [0.25cm] \hline 
%  Obscure       & Religious sect, founded Penn.: & Quaker  & Watery wisdom:  & Hydrology\\
  Obscure       & Carrier for your puppy: & Kennel  & Watery wisdom:  & Hydrology\\
 & \emph{Opacity}: $58.7^\circ$ & \emph{Obscurity}: 1.51 wpm  & \emph{Opacity}: $86.1^\circ$ &  \emph{Obscurity}: $<$ 1 wpm  \\
  \end{tabular} }
\end{center} 
\end{table}
\subsection{Obscurity as Word Frequency}

We operationalize obscurity by how frequently a word is encountered in common use. Such a measure has been shown to be a consistent predictor of performance on tasks such as word recall and discrimination~\shortcite{Bormann,Turner}. These effects have been explained, for example, by models that focus on the growth and structure of semantic networks. \citeA{Steyvers}, for instance, show that higher frequency words have greater connectivity of words that have low usage frequency.  

We use word frequencies from the Google Natural Language Corpus ~\shortcite{Michel2010} as a context-independent proxy for obscurity. This is an imperfect proxy in some applications, since obscurity may be context dependent (what is obscure at a caf\'{e} may not be in a machine shop), but is a natural choice for the datasets we use here, as both pose questions in a largely context-free setting.

\subsection{Opacity as Word2vec Distance}

We operationalize opacity using a machine-learning model of semantic content. Specifically, we use word2vec, a vector space model founded on the hypothesis that words that appear together and in the same context are likely to be semantically related~\cite{Baronietal2014a}. These relationships can be represented geometrically by mapping, or embedding, words into a continuous vector space where relative distance captures the richer notion of semantic relatedness. Word2vec is a model~\shortcite{Mikolovetal2013a} trained on a corpus of three million words and phrases from Google News articles, that places (``embeds'') words in a 300--dimensional space.

Word2vec relies on the assumption that words found nearby each other in a sentence tend to be related semantically. The words need not be exactly adjacent, hence the term ``skipgram'' for the class of models in which word2vec is found. A sentence discussing computers will preferentially contain words from that domain, for example, which allows word2vec to embed ``keyboard'' and ``trackpad'', for example, nearby each other the vector space. The embeddings can be surprisingly fine-grained, and capture relative positions: an early example is that computing the vector sum of ``king'', minus ``man'', plus ``woman'' puts one in the neighborhood of  ``queen''.
	
Since word2vec maps words to vectors, it allows us to quantify semantic relatedness using cosine similarity, a measure of closeness of two vectors equal to the angle between them. For example, given the vectors of two words, $x$ and $y$, the vector space representation provided by word2vec allows us to compute a measure of semantic similarity,
\begin{equation}
%             \textrm{cosine similarity} = \arccos{\left(\textrm{vec}(\textrm{x})\cdot{} \textrm{vec}(\textrm{y})/ ||(vec(\right)}
%             
	\textrm{cosine similarity} = \arccos{\left(\frac{\vec{x}\cdot\vec{y}}{\|x\|\|y\|}\right)}
    \label{vec_example}
\end{equation}
that corresponds to the angle between the directions of the two words. If, for example, we compute the cosine similarity for ``breakfast'' and ``bagel'',  the calculation yields a relatively small angle, $66^\circ$, indicating that they are reasonably related to each other.\footnote{The high-dimensionality of the word2vec model, and the sparseness of the space, mean that small angles (say, $10^\circ$ or lower) are extremely rare. An angle of $66^\circ$ indicates anomalously high alignment;  it is out on the tail of the word-to-word angle distribution, and well outside the null.} Two semantically unrelated words, such as ``breakfast'' and ``torus'', are more orthogonal (in this case, nearly perfectly orthogonal, with an angle of $88^\circ$). These simple examples show how cosine similarity can provide a measure of semantic relatedness, and results like these are a source of much of the interest in the underlying skip-gram model of semantics provided by word2vec.

Eq.~\ref{vec_example} measures the extent to which latent cues within a question function either independently or in synergistic combination. Consider the crossword clue ``tasty torus'' and its answer, ``bagel''. If a clue signals an answer through a combination of all clue words, the solver should consider the clue, ``tasty torus'', in its entirety. If, however, the information value of the clue is provided by the words in an independent fashion, the solver can attend to each word, ``tasty'' and ``torus'' in turn. 

These intuitions are captured by two distinct models (for a discussion of a third model, ``Keyword'', in which we find the cosine similarity between an answer word and the maximally similar clue word, see Appendix). The ``Synergistic'' case compares the whole clue to the answer by calculating the cosine similarity between an answer word and the sum of all clue words. The ``Independent'' model averages the similarity between answer and clue words taken in turn. The two models differ because the synergistic case includes the relative vector lengths, rather than just the average differences between the different directions.
    
\begin{table}[b]
\begin{center} 
\caption{Example cosine similarity calculations in the Synergistic and Independent model for a question-answer pair, in which $A$ is the answer, and $Q1$ and $Q2$ are question words.} 
\label{Table2} 
\begin{tabular}{l|l} 

%Model & \\ [0.5cm]
%\hline

Synergistic      	& $\frac{\xshlongvec[1]{A}\cdot \left(\xshlongvec[1]{Q1}+\xshlongvec[1]{Q2}\right)}{\|A\|\|\xshlongvec[1]{Q1} + \xshlongvec[1]{Q2}\|}$ \\ [.5cm]
Independent      &  $\left(\frac{\xshlongvec[1]{A}\cdot\xshlongvec[1]{Q1}}{\|A\|\|Q1\|} + \frac{\xshlongvec[1]{A}\cdot\xshlongvec[1]{Q2}}{\|A\|\|Q2\|}\right) / 2$ \\

%\hline
\end{tabular} 
\end{center} 
\end{table}

\subsection{Data sets}

%%% TK TK what was the original size?
Our first corpus is a set of 677,512 clue--answer pairs from \emph{New York Times} (NYT) crossword puzzles published daily between November 22, 1993 and July 12, 2017. We exclude all non-word clues (e.g. ``!'', ``9:59''), and clues that explicitly indicate that the answer is an abbreviation. For the remaining clue and answer pairs, we exclude stopwords (such as ``and'', or ``the'') and keep only nouns, verbs, and adjectives. Finally, any clue and answer pair for which the answer word and at least one clue word does not appear in the vocabulary of our word2vec model is excluded, eliminating non-words and compounds. Our resulting data set has 471,342 clue and answer pairs. The answer in this subset is always one word, and the average length of the clue is 2.9 words.

The set of Jeopardy questions we use consists of 216,930 questions aired on the T.V.\ show between 1984 and 2017. We refine our set to include only nouns, verbs, and adjectives. As in the case of the crossword puzzles, Jeopardy answers can be more than one word but for simplicity we constrain our set to those with one-word answers. Our final set contains 104,293 question-answer pairs. The average clue length in our set is 7.9 words.

Puzzles in both datasets vary in difficulty in a systematic and marked fashion. In the case of the NYT crosswords, the easiest puzzles are published on Monday and ascend in difficulty, day by day, to the hardest puzzles which appear on Saturday (Sunday's crosswords have the same difficulty rating as Friday's). We can thus tag all clue and answer pairs for level of difficulty (1 = easiest, 6 = hardest) according the day of the week they appeared. Historically, the difficulty level was approximated by puzzle setter through intuition~\cite{almen}; more recently, these levels have been validated through user analysis through the mobile application Puzzazz~\cite{puzz}.

In Jeopardy, within each category (say, ``history''), questions are given a monetary value that serves as a proxy for difficulty. In regular Jeopardy, questions are valued between 200 and 1000 USD, and in Double Jeopardy, these values are doubled; we collapse across these two rounds, marking questions by monetary value only. The question set we use contains both game types and for convenience and consistency, we refer to 200USD questions as difficulty 1 and 2000USD, as difficulty 8. A study of performance on Jeopardy questions by value conducted by~\citeA{casino} shows that value is a strong predictor for difficulty. Over thirty-three seasons of the show, questions at the current minimum 200USD level were answered at a average rate of 6.3 correct answers to every incorrect answer (6.3:1), while questions at the 2000USD level were answered at an average rate of 0.8 correct answers for every incorrect one, nearly a factor of ten lower.

A property unique to the NYT crosswords provides a useful check of our methods. Clues have a consistent typology which includes a type of word play or pun clue, demarcated by the use of a ``?''. To answer a word-play clue such as ``Make a good impression?'', a successful solver has to both recognize the figurative expression and realize that the impression meant here is a literal carving. In contrast, a non-pun clue ``Engrave glass with acid'' points the solver directly to a definition so all they must do is consider where this answer would fall on the grid, and retrieve the word ``etch''. We use this to making to classify  pairs as either non-pun (454,821 pairs) or pun clues (16,521 pairs); the nature of word-play means we expect it to have greater levels of opacity, which ought to be reproduced in our data. At the same time, the marking allows us to exclude these unusual clues, eliminating a difference and making the comparison between our sets more direct.

An additional sense check is provided by a simple null model: we take the average for the data set of all clue and answer pairs, puns, non-puns, and compare to a null model in which we randomly match an answer with a clue. This comparison allows us to check that in our method, clues and answers are more aligned than would be by chance. 

\section{Results}

We consider obscurity and opacity in turn, showing how these independent measures drive question difficulty. Our data is sufficiently rich that we can measure the covariance of difficulty with one measure while controlling for the other, allowing us to detect potentially non-linear relationships between the two dimensions. At the end, we turn to a simpler regression model to explore the possibility of additional dimensions over and above the two that appear to dominate.

\subsection{Obscurity}
Jeopardy questions show strong differences in obscurity as a function of difficulty: easier questions have answers far more common in ordinary use (median 11.22 per million words) than difficult ones (median 6.61 per million words). That is, they vary obscurity in the expected direction as part of a strategy for increasing overall difficulty. 

\begin{figure}[t]
\begin{center}
\includegraphics[width=1\textwidth]{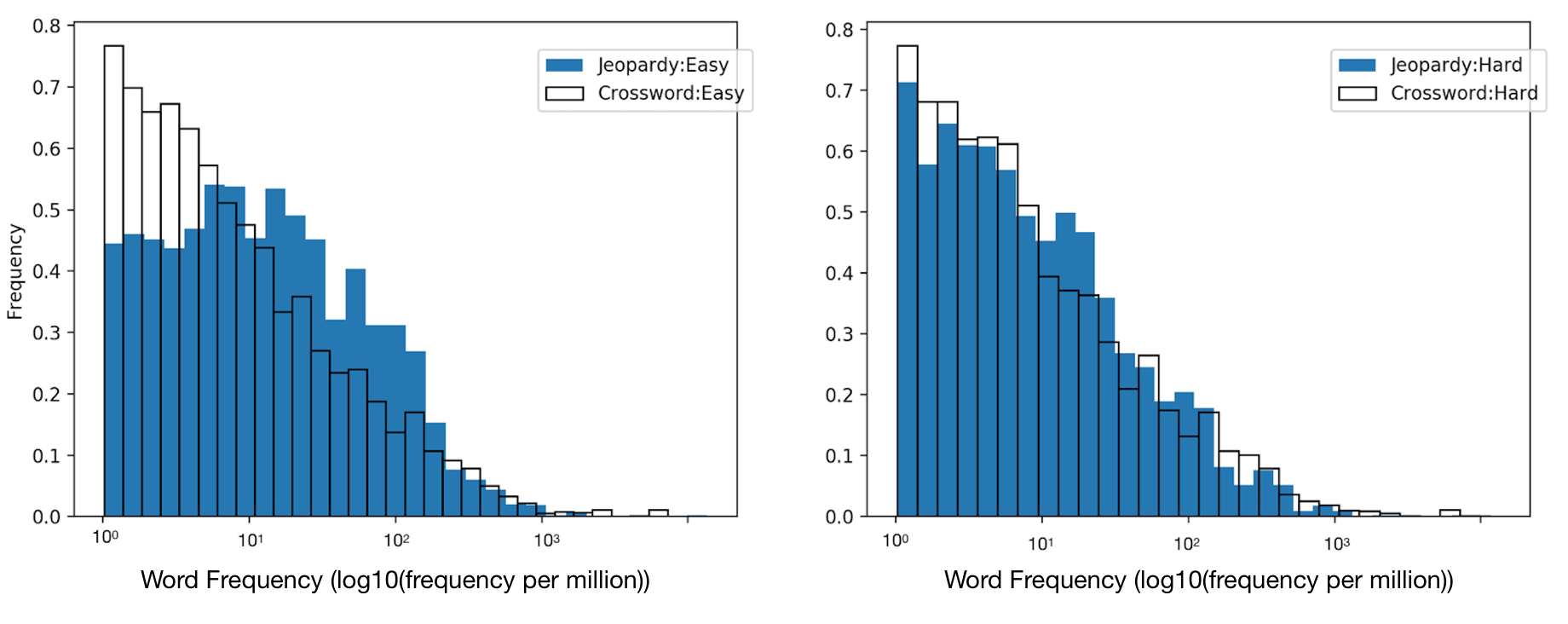}
\end{center}
\caption{Word frequencies of easy and difficult Jeopardy and crossword answers. The $x$-axis scale is $\log_{10}{\textrm{(frequency per million words)}}$. While easy and difficult crossword clues have reasonably similar word-frequency distributions, easy Jeopardy questions tend to have answers drawn preferentially from more common words as compared to difficult questions. } 
\label{wordFreq}
\end{figure}

Crossword clues, by contrast, show at best only weak differences in word frequency between easy and hard (Fig.~\ref{wordFreq}). The median frequencies for hard vs.\ easy questions (5.62 and 5.63 per million words, respectively) are not significantly different ($p>0.1$).

\subsection{Opacity}

Our operationalization of opacity passes simple sanity checks. Crossword clues align with answers more closely than would be expected by chance, as do Jeopardy questions and corresponding answers; both at high significance (as expected, given the amount of data to hand; see Appendix). The unique word-play property of crosswords provides a separate check: non-puns are more aligned with their answers than puns (indicated by a terminal ``?''; see Appendix). 

\begin{figure}[!t]
\begin{center}
\includegraphics[width=0.65\textwidth]{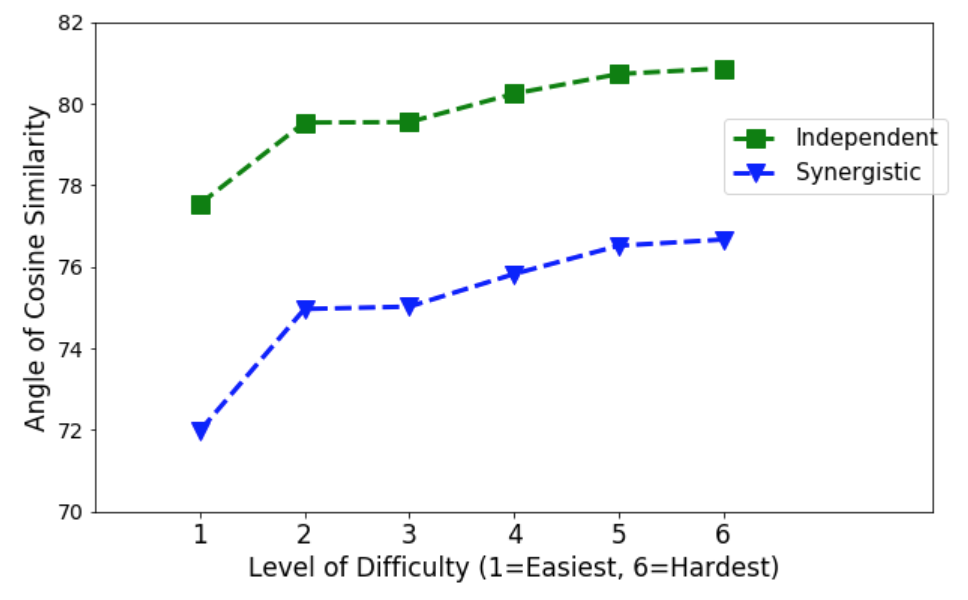}
\end{center}
\caption{Average cosine similarities in Synergistic and Independent by question difficulty in crosswords. Clues align less closely with answers in the Independent than in the Synergistic case. Indirection increases with difficulty; at the same time, the boost from synergistic combination (the gap between the blue and green lines) decreases. Independent samples t--tests demonstrate significant differences between the Independent and Synergistic models at $p\ll 10^{-3}$; bootstrapped standard errors are smaller than symbol sizes.} 
\label{opacity}
\end{figure}

We consider the case of crosswords first. A comparison between the Synergistic and Independent models shows the extent to which the value of a clue is found in a combination of all words, or contained primarily in individual words (Figure~\ref{opacity}). Overall, opacity increases as clues become more difficult: hard clues are less aligned with their answers than easy clues. Additionally, it is generally the case that synergy works: the average cosine similarity is greater in the Synergistic than in Independent model by $4.6^\circ$. This suggests the puzzle-solver who can attend to how words combine has an easier time than one who considers each word in turn. 

Increased difficulty frustrates the synergistic strategy: the benefit of a synergistic combination of words declines as difficulty increases. While more difficult clues have higher rates of indirection, their synergistic angles increase more quickly, approaching the independent case. This can be seen in the narrowing gap between the green and blue lines in Figure 2: while synergy gains the easy-puzzle solver $5.6^\circ$, it gains the difficult-puzzle only $4.2^\circ$ . This $25\%$ decrease in returns to synergy, while modest, is distinguishable at the $p\ll 10^{-3}$ level.

In crosswords, the role played by opacity is sufficiently strong that it can be seen directly in its effect on difficulty. Understanding the more general relationship between obscurity and opacity requires that we see how difficulty covaries with one while we control for the other.

%TKTK
%, and is the primary axis along which questions are made difficult. WE find that difficult q are not only more indirect than easy clues, but they also have limited gain from synergistic combination of words.

\begin{figure}[t!]
\begin{center}
\includegraphics[width=0.75\textwidth]{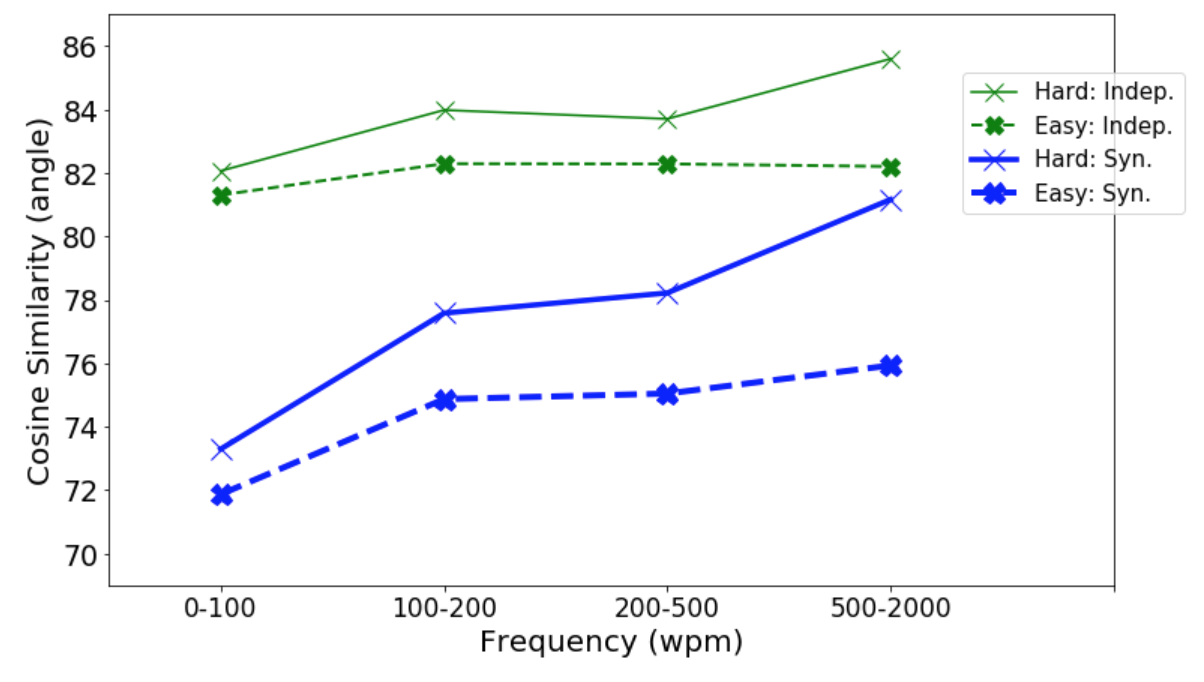}
\end{center}
\caption{Cosine similarities, as a function of word frequency. We plot separately, easy (dashed lines) and hard (solid lines) Jeopardy questions, as well as the cosine similarities for the independent (green) and synergistic (blue) models, and four bins of word frequency. We see that for all frequencies and in both models, hard questions are more opaque than easy questions. This difference is, however, greater when answer words are more common, than when they have low frequency of occurrence.} 
\label{jep_opacity_freq}
\end{figure}

\begin{figure}[t!]
\begin{center}
\includegraphics[width=0.99\textwidth]{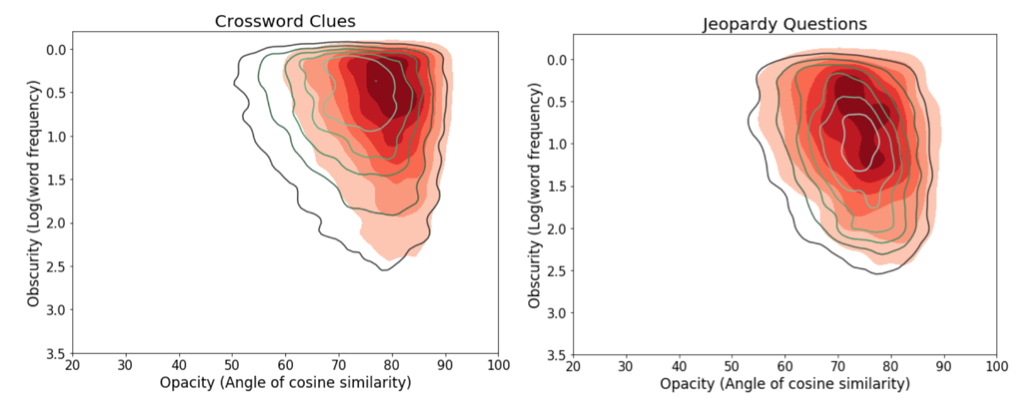}
\end{center}
\caption{Kernel Density Estimations of the distribution of opacity and obscurity in both hard (shaded contours) and easy (outlined contours) crossword and Jeopardy questions. Opacity is cosine similarity as calculated by our Synergistic model and obscurity is $\log_{10}{\textrm{(frequency per million words)}}$, with lower frequency values at the top). Contours reflect the topology of the distribution, with the single central region of each distribution containing the highest density of data. A horizontal line from the left (\emph{i.e.}, controlling for obscurity), or a vertical line from below (\emph{i.e.}, controlling for opacity) intersects the open contours first.}  
\label{kde}
\end{figure}

In the case of Jeopardy, the bias towards obscurity hides the role of opacity: once we control for the difficulty induced by shifting obscurity, the compensatory roles that opacity and obscurity play emerges. This is shown in Fig.~\ref{jep_opacity_freq}, where within a given level of frequency, difficult Jeopardy questions are more indirect than easy questions in both the Synergistic and Independent models. The full distributions are shown in Fig.~\ref{kde}; along any horizontal or vertical line, the harder questions are shifted towards greater opacity, or greater obscurity, respectively. While the relationships in both cases are non-linear, they are monotonic in difficulty.

Control for obscurity also recovers the diminishing returns to synergy as difficulty increases: the gap between the solid lines (in Fig.~\ref{jep_opacity_freq}, hard question independent and synergistic indirection) is narrower than that for the easy case. While a synergistic combination of words awards a question answerer $9.44^\circ$ of increased alignment in easier, but infrequent, answers, they gain only $8.77^\circ$ for difficult questions. This  benefit of synergy  becomes increasingly frustrated when the answer words are the most common; here, synergy aids the solver of easy questions by $6.27^\circ$ but those of hard questions, only $4.43^\circ$.

\subsection{Beyond Opacity: Additional Dimensions of Difficulty}
\label{additional}
%%%%%%%%%%%% UPDATED TABLE CAPTION
\begin{table}[!t]
\begin{center} 
\caption{Regression results predicting standardized ($z$-score transformed) difficulty such that increasing  values represent harder questions. All predictors are similarly standardized. Obscurity is $-\log_{10}{\textrm{(frequency per million words)}}$ such as obscurity increases, the answer word becomes less frequent and opacity is the degree of cosine similarity as calculated by our Synergistic model. In Model III we include the density of the semantic network surrounding the answer word in our word2vec model. We estimate this as the inverse of the euclidean distance between the answer and the word that is closest but shares fewer that 90\% of the same letters. This latter restriction is imposed to exclude word stems. In Model IV we include other predictors, beyond just obscurity and opacity: the number of words in the question, the frequency of the least common question word (in $\log_{10}{\textrm{(frequency per million words)}}$), and the frequency of conjunctions (e.g. ``and'', ``or'') in the question. Standard errors are reported in parentheses, and *** $p\ll 0.0001$, ** $p= 0.01$, and * $p= 0.05$.} 
\label{reg}
\resizebox{\textwidth}{!}{%
\begin{tabular}{lllllllll}
 \multicolumn{9}{c}{Dependent Variable: Question Difficulty} \\ [0.2cm]
      & \multicolumn{4}{c}{Crosswords} &  \multicolumn{4}{c}{Jeopardy} \\ 
      & (I) &(II) & (III) & (IV)  &(I) & (II) & (III) & (IV) \\ [0.2cm]\hline
  	Obscurity &0.023*** &0.038*** & 0.041***  &0.044***		& 0.068*** & 0.070*** & 0.078*** & 0.076***\\
    	& \emph{(0.002)} &\emph{(0.002)} & \emph{(0.002)} &\emph{(0.002)}  		& \emph{(0.005)} & \emph{(0.005)} & \emph{(0.005)} &\emph{(0.005)}\\[0.2cm]
  	Opacity & & 0.149***	& 0.150***	& 0.149*** 		& &0.059***& 0.052*** &0.053*** \\ 
        	&  &\emph{(0.002)} & \emph{(0.002)} & \emph{(0.002)} 		& & \emph{(0.004)}& \emph{(0.005)} &\emph{(0.005)}\\  [0.2cm] 
    Ans. density & & & 0.025*** & 0.027***		& & & 0.057*** &  0.057***\\ 
    		& & & \emph{(0.004)} & \emph{(0.004)} & & & \emph{(0.009)} & \emph{(0.009)} \\ [0.2cm]
    Q. length & & & & 0.003*** & & & &0.012*** \\
    		& & & & \emph{(0.002)}& &  & &\emph{(0.002)} \\  [0.2cm] 
    Min. Q. & & & & 0.027*** & & & & 0.012*\\
    word freq.	& &	& & \emph{(0.002)} & & & & \emph{(0.005)}\\  [0.2cm] 
    Freq. of conjunctions & & & &-0.029*** & & & &0.0065\\
     in Q.	& &	& & \emph{(0.002)}& & & & \emph{(0.005)}\\  [0.2cm] \hline
    		
  	N       & 278,497	& & &		& 53,928 & & & \\ 
 	$R^2$	& 0.001 & 0.023 	& 0.023& 0.024	&0.005	& 0.007 & 0.008 & 0.010\\ 
    AIC		& $7.90\times10^{5}$ & $7.84\times10^{5}$ & $7.84\times10^{5}$ & $7.83\times10^{5}$ & $1.21\times10^{5}$ & $1.208\times10^{5}$ & $1.208\times10^{5}$ & $1.207\times10^{5}$ \\
\end{tabular} }
\end{center} 
\end{table}

In both crosswords and Jeopardy, opacity and obscurity covary. The effect is weak, however: crosswords have $r = 0.097, p < 10^{-4}$; Jeopardy, $r = 0.194, p<10^{-4}$. Opacity and obscurity, in other words, can be orthogonal directions along which difficulty can vary. Though the effects are weakly non-linear, a regression analysis (Table~\ref{reg}) confirms this picture of titration between obscurity and opacity; inclusion of both obscurity and opacity yield a better fit model than obscurity alone. For both crosswords and Jeopardy, increases in frequency of the answer word result in decreases in difficulty and easier questions, while increases in angle of cosine similarity increase difficulty. The relative effects are not, however, equal: obscurity is more instrumental in Jeopardy, while opacity has the larger influence on crosswords. Jeopardy is more a test of general knowledge, while crosswords challenge a reader to traverse distinct conceptual spaces.

Opacity and obscurity are strong effects that generalize across the two datasets, but they do not exhaust the dimensions of question-asking difficulty. In our third regression model we include the density of the semantic network surrounding an answer word, as generated by our word2vec model.\footnote{We thank Danny Oppenheimer for this suggestion.} In both cases, as the density of words surrounding the answer increases, difficulty increases. Words that lie in denser areas are harder to discriminate among: the process of choosing the one correct answer is more difficult. 

%increased network density around the answer predicts a decrease in difficulty: words that lie in areas more sparse are more difficult to retrieve and select. 

The addition of additional predictors, largely orthogonal to both opacity and obscurity, and favored by AIC, increases our ability to predict difficulty. 

In particular, longer questions predict increased difficulty in both data-sets, suggesting a role for some measure of cognitive complexity. Conversely, the frequency of conjunctions in the question (``and'', ``or'', ``but'', and so forth) predicts reduced difficulty, suggesting that these questions can compensate for the difficulty of longer questions by providing multiple cues to aid the answerer. Finally, the presence of obscure words in the question (as opposed to the answer) predicts a decrease in difficulty: rare words may provide a detectable high-value signal. These latter two signals, however, only appear strongly in the crossword case, suggesting that they may be subordinate effects that are more pronounced when questions are varied perpendicular to the obscurity axis.

\section{Discussion}

%% Our semantic spaces in this case are public, in the sense that the structure in question is independent of person answering; we

What makes a question difficult is more than the rarity of its answer. Our results demonstrate the existence of opacity, a separate, mostly orthogonal axis of difficulty. Clarity, or directness of the relationship between question and answer, plays a key role that may be of equal, or greater, magnitude than that of obscurity. Both dimensions are found in two distinct datasets, with different rules and asking and answering constraints.

Obscure answers are hard to find because they are infrequently encountered in ordinary life. Simple Bayesian models can capture this dimension as a process of overcoming low priors~\cite{chater}, or as a search problem made harder when the question-answerer must explore less-encountered conceptual spaces. Such models need, at least, to be extended to account for how the contextual features of question cues facilitate or hinder arrival at the answer. The role of interpretation and semantic ambiguity suggests that Bayesian models need to allow priors to depend on the domain inferred from the question words, in such a way that re-weights degrees of belief on the basis of distance from the domain region. Even this extension may be insufficient, however. In the case of Jeopardy the correct domain of the answer is specified by the question category, and yet opacity still plays a role in the question's difficulty.

These results suggest that features of the cognitive process of question-answering plays a role in making questions difficult.  \citeA{Turney2010} has shown that the neural network based approach we chose as an operationalization of this dimension yields representations that are highly correlated with human judgments of semantic similarity. \cite{Griffiths2007} show that other classes of distributional semantic models have been shown to have cognitive plausibility. Both results raise the likelihood that models like word2vec, that build accounts of meaning through co-occurrence, can capture ways in which we process language. The effect of opacity on question difficulty may be best described by a process model that describes how individuals navigate the semantic cues within a question. A question that employs words from vastly different domains (such as the earlier example of ``iron horse'') will yield a more orthogonal angle of opacity, reflecting the intuition that what makes that question difficult is the use of context-inconsistent words that makes the path to ``train'' unclear.

These process models provide an complementary angle on the literature that conceives of questions as programs that operate on the world to produce an answer~\cite{CohenLake2016a}. Opacity can, with some effort, be reduced by, for example, becoming familiar with the language of the domain in question before asking the question. Obscurity may be more difficult, but still not impossible, to change: if I believe the answer I'm seeking may be too obscure for my interlocutor, I might redirect, seeking a simpler answer at the cost of some utility. Opacity and obscurity are expected play a role in the formulation of rational intentions and goals in question-asking and in dialogue~\shortcite{Hawkins2015}. 

We find that the dimensions of obscurity and opacity can also compensate for each other. While the most difficult questions may lie at the upper bounds of both opacity and obscurity, within our test set of successful puzzles, question askers can trade opacity and obscurity off of each other, tempering difficulty arising from indirection by making the answer a more common word.  That the opacity-obscurity tradeoff can be found in two distinct practices suggests the possibility that these are truly generalizable features of question asking, showing how Indirection and contextual effects alter the transmission of information and the ways in which concepts are conveyed through language.

We have focused on establishing opacity as new a dimension along which difficulty increases, or conversely, could be reduced by rephrasing the question. Other dimensions are also in play, although to a lesser extent in our data. Our regression results suggest a role for one type of complexity, longer questions are more difficult. This argues against Bayesian accounts that predict that increasing the number of words in a question increases the amount of available information, thus decreasing its difficulty. A process model may help explain our contrary result by reference to the costs of processing. 

Another way in which questions can become easier, as revealed by our results, is the distribution of information throughout the question. Easier questions tend to have a low-frequency word that that may play a signaling role. Conjunctions also reduce question difficulty, perhaps because they provide intersecting clues that enable answers to narrow down the space even with partial knowledge.

These additional effects bear on question-asking models that focus on the computational aspects of asking and answering. \citeA{Rotheetal2017a} offer a different angle on question composition in terms of an information-theoretic measure, Expected Information Gain (EIG), where the value of a question is the expected reduction in uncertainty about a hypothesis, averaged over the set of all possible answers. Questions with high EIG scores provide more information about the true answer, while questions with low scores reduce uncertainty less. In constructing questions to maximize EIG, a question asker may favor increasingly complex questions; our results suggest that while these answers may be more valuable, they also might be harder to provide.

%%%%%%
Our results suggest testable hypotheses for studies of question-asking beyond the domain we consider here. However, not all the questions we encounter in life have the necessary structure for these independent variables to apply. To help predict which kinds of questions might follow the patterns we find, we propose a classification scheme for separating questions. Our simple scheme draws on a pair of distinctions, and thus a classification of question types into four distinct categories (Table~\ref{typesQ}).

The first distinction is whether the answers are retrieved or constructed. The questions considered here require retrieval: recalling the steps of a recipe, or the name of someone you recently met. Alternatively, questions may require the construction of an answer: the best route between a pair of cities (when the pair has not previously come to mind), or the choice of a locker combination.

In addition, the answers themselves may be located in a pre-existing structured space: in a recipe for custard, some choices (adding sugar, or milk) are `closer' to the space where custard-like recipes live, while other choices (such as adding ketchup, or braising with butter) are further away. Or, they may be found in a space without any notion of near or far: there are many names for an acquaintance, but none of them are likely to be more appropriate for the person in question; there is no sense in which the decisions of which combination to use is closer to the correct answer. The boundaries between different question types is not firm, of course. Constructing a path between two cities may involve recalling a path previously taken from a city center to the highway, and the space of names has some structure dictated by, for example, gender.

%%% put in 2x2

\begin{table}[!b]
\begin{center} 
\caption{Classification of question types. This work studies questions of the structured-retrieved type.} 
\label{typesQ} 
\begin{tabular}{l|ll} 

      &  \emph{Structured}  & Unstructured \\ \hline
Constructed       	&  How do I drive to Pittsburgh  & What password should  \\
& from Santa Fe? & I choose?\\ [0.25cm] \hline
\emph{Retrieved} & What goes well with leeks? & What is his name? \\

\end{tabular} 
\end{center} 
\end{table}

While we believe that our results are most closely applicable to the structured-retrieval case, three features of our datasets make them different from question asking in the wild and provide challenges for direct generalization. First, all of the questions we consider here are ``successful'', or fair: sufficiently well-framed that, no matter how difficult, they can be seen, at least in retrospect, to point to the correct answer. While this enables us to focus on dimensions such as opacity, obscurity, and complexity, it leaves out ambiguity, one of the most basic ways in which sentence comprehension and question-answering can be hard \cite{Church1982}. A full empirical account of difficulty needs to measure question-asking along this axis.

Second, our questions ask about public, rather than subject-relative, information. Questions such as``what did you have for lunch yesterday'', or ``when did you last visit the dentist'' differ from questions in our data in a number of ways, but are common in everyday life. It may be possible to extend our notion of obscurity and opacity to these cases. Personal memories are themselves organized in a structured fashion, and personal questions can either conform or deviate from the structures that hold the memories in question. ``What did you eat for lunch'' may be a less opaque question than ``what did you read during lunch'', for example, if memories of lunchtime are organized around food, rather than what one reads. In the absence of other forms of data, of course, such extensions are necessarily speculative.

Third, our questions are framed by someone who already has the answer in mind; difficulty is adjusted in a voluntary fashion by the question asker. Real-world validations (see Methods) show that the difficulty they create is real, but our data can not tell us about the degree to which opacity and obscurity play a role in the difficulty of answering questions in real world scenarios.

We suggest two investigations to overcome the limitations of this third feature. A number of websites online, such as Quora, Stackexchange, and Reddit, provide forums for users to pose questions. Our operationalizations here can be applied directly to question-answer pairs to measure obscurity, opacity, and the additional measures of Sec.~\ref{additional}. Meanwhile, the delay between the posing of a question and its answer appropriately controlled for latent variables, may provide an approximate measure of difficulty, Stackexchange provides a particularly nice venue for this kind of work, because the best answers to a question are rated and marked by other users, and strong norms exist against digression and unnecessary content.

Laboratory studies may overcome this limitation in a different fashion. A subject incentivized to gain information from other participants could provide a source of questions, while the timing of response, the number of requests for clarification, and the fraction of correct answers could provide a measure of difficulty. The challenge is to operationalize obscurity and opacity in such a way as to overcome the smaller amount of data expected from a lab-based study. The Google Natural Language corpus, and word2vec, are both trained on cross-domain sets, and may have insufficient signal for studies of question/answer pairs in the hundreds or thousands, rather than hundreds of thousands. More precise semantic networks may be generated through word-association tasks by participants on the same tasks, for example.

\section{Conclusion}

The difficulty of a question can be found, in part, in the relationship it holds to its answer. Our analysis of crosswords and Jeopardy questions reveals new dimensions to this basic relationship. As the angle between the answer and the words used to phrase the question grows, the path to the answer becomes increasingly indirect and the possible spaces of answers harder to narrow down. Such a process is further complicated by the rarity of the final answer within these spaces. Knowing a fact is not enough: it is necessary to find a path towards it.

\section{Acknowledgements}

We thank Gretchen Chapman, Danny Oppenheimer, George Lowenstein, Russell Golman, Stephen Broomell, Nik Gurney, and participants in the Fall 2017 Large-Scale Social Phenomena Graduate Seminar for feedback on early drafts of this work. We thank Christopher Bates, Shane Mueller, and George Kachergis for useful suggestions. This research did not receive any specific grant from funding agencies in the public, commercial, or not-for-profit sectors.

\bibliographystyle{apacite}

\clearpage

\section{Appendix}

\subsection{Keyword Model}

%\begin{table}[b]
%\label{keyword_tbl}
%\begin{center} 
%\caption{Keyword model in which only the question word most aligned with the answer is considered. Here, tasty is more aligned with bagel than torus so the Keyword model finds the cosine similarity between bagel and tasty.} 
%\begin{tabular}{ll} 
%\hline
%Model & \\
%\hline
%Keyword     &    $\textrm{max$\{$}{\textrm{vec(bagel)}\cdot{}\textrm{vec(tasty)}, \textrm{vec(bagel)}\cdot{}\textrm{vec(torus)}}$\}$$  \\
%\hline
%\end{tabular} 
%\end{center} 

%\end{table}
% max{vec(bagel)*vec(tasty), vec(bagel)*vec(torus)}
In the main text we presented two models of opacity -- ``Independent'' and ``Synergistic''.  In our analysis we also considered a third model, ``Keyword'', in which we found the cosine similarity between the answer and the maximally similar clue word. This case is a \emph{post hoc} measurement that does not capture the underlying challenge facing the interlocutor, but does provide a sense of the extent to which an answer, when found, was concentrated in one place versus another. We find that in both crosswords and Jeopardy, once the clue word with the most information has been identified, that one word is more closely aligned with the answer than either of our other two models. 

\subsection{Null models}

As checks of our opacity measure, we construct a null models for both crosswords and Jeopardy, and for each of our ``Synergistic'' and ``Independent'' models. In Table~\ref{xwrods_appendix} and Table~\ref{jep_appendix}, we present the average cosine similarities for each model. Importantly, crossword clues align more closely with their answers than would be expected by the chance, as do Jeopardy questions with their answers as well. 

An additional check is provided by the puns in our crossword data (Table~\ref{xwrods_appendix}). In both the ``Synergistic'' and ``Independent'' models, the non-puns are more aligned with their answers than the puns. For example, given the same answer of ``Etch'' the non-pun clue ``Engrave glass with acid'' is more aligned with the answer than the pun clue, ``To make a good impression''.

%%%%%% table
\begin{table}[b]
\begin{center} 
\caption{Average cosine similarity (angle) for each data category in crosswords. The ~\emph{Data} category includes all clue and answer pairs. Independent samples t--tests demonstrate significant differences between data, null at $p\ll 10^{-3}$.} 
\label{xwrods_appendix} 
\vskip 0.12in
\begin{tabular}{llll} 
\hline
    &  Synergistic & Independent  \\
\hline
Data       	&   75.0* & 79.6*	\\
Puns   		&   78.8*	& 81.9* 	\\
Non-Pun     &   74.9* &	79.5*	\\
Diff. 1 (Easiest)    &   71.9*	& 77.5*	 \\
Diff. 6 (Hardest)		& 	76.7*	& 80.9*	\\
Null		& 	84.4	& 85.7\\
\hline
\end{tabular} 
\end{center} 
\end{table}

%%%%%% table
\begin{table}
\begin{center} 
\caption{Average cosine similarity (angle) for each data category in Jeopardy. Independent samples t--tests demonstrate significant differences between data, null at $p\ll 10^{-3}$.} 
\label{jep_appendix} 
\begin{tabular}{llll} 
\hline
  &  Synergistic & Independent  \\
\hline
Data       	&   71.8* &	81.3* \\
Diff. 1 (Easiest)   	&   71.8* &	81.4* \\
Diff. 8 (Hardest) 	&   71.6* &	81.2* \\
Null	   	&   81.0 &	85.6 \\
\hline
\end{tabular} 
\end{center} 
\end{table}

\clearpage

\setlength{\bibleftmargin}{.125in}
\setlength{\bibindent}{-\bibleftmargin}

\bibliography{mainText}

\end{document}